\documentclass[a4paper]{article}
\pdfoutput=1
\usepackage{amsmath,amsfonts,graphicx,bm,subfigure}
\usepackage{verbatim}

\usepackage{INTERSPEECH2019}

\title{Semi-supervised voice conversion with amortized variational inference}
\name{Cory Stephenson, Gokce Keskin, Anil Thomas, Oguz H. Elibol}
\address{Intel AI Lab, Santa Clara, CA, USA}
\email{cory.stephenson@intel.com}

\begin{document}

\maketitle
\ninept
\begin{abstract}
In this work we introduce a semi-supervised approach to the voice conversion problem, in which speech from a source speaker is converted into speech of a target speaker. The proposed method makes use of both parallel and non-parallel utterances from the source and target simultaneously during training. This approach can be used to extend existing parallel data voice conversion systems such that they can be trained with semi-supervision. We show that incorporating semi-supervision improves the voice conversion performance compared to fully supervised training when the number of parallel utterances is limited as in many practical applications. Additionally, we find that increasing the number non-parallel utterances used in training continues to improve performance when the amount of parallel training data is held constant.
\end{abstract}


\section{Introduction}
\label{sec:intro}
The goal of voice conversion (VC) is to take in speech produced by one person (source) and process it such that it sounds like it was produced by a different (target) speaker. VC systems have a diverse set of potential applications including the construction of more natural synthetic voices, anonymous transmission of recorded speech, voice spoofing, and data normalization for further speech processing applications. Due the broad applicability and inherent difficulty of the problem, the design of VC systems has been an area of consistent interest for decades and continues to see active research \cite{toda2016voice, lorenzo2018voice}.

Of particular interest are the statistical approaches that do not require access to word or phonetic transcriptions. These include the early work using vector quantization \cite{abe1990voice}, which inspired the popular Gaussian mixture model (GMM) conversion method \cite{stylianou1998continuous} and its later improvements \cite{toda2007voice}. While these methods can produce good results, they require many parallel utterances for training, are sensitive to misalignment in the training data, and special care must be taken to avoid the 'buzzing' that arises due to the usual maximum-likelihood training objective \cite{toda2007voice}.

The necessity of parallel data for training is quite limiting, as the collection of this kind of data is a slow and expensive process. The desire to avoid this requirement has led to the development of VC approaches which use only non-parallel data from the source and target speakers \cite{erro2010inca, mouchtaris2004non, kinnunen2017non}. While removing the parallel data requirement eases the burden on data collection, it introduces extra difficulty in ensuring the converted speech is both high quality and unchanged in phonetic content \cite{saito2018non}.

More recently, advances in training deep nonlinear models have led to renewed interest in applying neural network techniques to the VC problem. Methods which use deep neural networks as feature extractors or to parameterize the conditional distributions in generative models have proven effective in doing VC both with \cite{desai2009voice, sun2015voice, nakashika2013voice, kaneko2017sequence} and without \cite{hsu2016voice} parallel data. These models excel at producing natural sounding speech. However, the increased model complexity comes with an increased demand for larger quantities of data in both the parallel and non-parallel case.

It is the goal of this work to fill the gap between parallel and non-parallel data voice conversion by introducing a method that uses both types of data simultaneously during training. To do this, we frame voice conversion as a semi-supervised learning problem. This follows naturally from previous shallow generative approaches such as \cite{stylianou1998continuous, toda2007voice}, which make use of a shared set of latent variables that generate both the source and target speech. While coupling this type of model with nonlinear transformations of the latent variables yields an intractable inference problem, we find that amortized variational inference as applied to deep generative models \cite{DBLP:journals/corr/KingmaW13, kingma2014semi} makes both training and conversion efficient. 

We demonstrate the effectiveness of semi-supervised training in multiple ways.  First, we extend a well-known neural network VC algorithm \cite{sun2015voice} such that it can be trained with semi-supervision.  Then we show that incorporating non-parallel data in training leads to higher quality voice conversion when parallel data is scarce.  Additionally, we verify that the semi-supervised training gives equivalent or better results to training with only parallel data when non-parallel data is scarce.  Finally, we confirm that conversion accuracy continues to increase with increasing amounts of non-parallel training data, albeit with diminishing returns. We also observe that the semi-supervised training results in audio of equal or higher perceptual quality than the parallel data conversion systems.

\section{Related work}
\label{sec:related}
While there are many proposed voice conversion methods in the literature (for a more in depth review, see \cite{mohammadi2017overview}), here we examine some of the popular statistical approaches which use parallel training data. A common way to think about this problem is to posit the existence of a latent variable $\bm{z}_t$ that describes the sound is produced at time $t$, but not the characteristics of the source that produced it. The job of the voice conversion system then is to \textit{infer} $\bm{z}_t$ from a sound produced by the source, and \textit{generate} the corresponding sound in the target's voice. 

\subsection{Gaussian mixture model voice conversion}
At a high level, this is how the well known GMM VC system \cite{stylianou1998continuous} achieves conversion. In this approach, the input consists of a sequence of features $\bm{X}=\{\bm{x}_1, \bm{x}_2,...,\bm{x}_T\}$ which are each assumed to be independently Gaussian mixture distributed. The means of the mixture components are parameterized with $\bm{\mu}_i$ covariance matrices with $\bm{\Sigma}_i$. In this model, $\bm{z}_t$ is a categorical latent variable that indicates which Gaussian is responsible for the observation. By virtue of the Gaussian mixture assumption, it is possible to compute the posterior distribution $P(\bm{z}_t = i|\bm{x}_t)$ for $\bm{z}_t$ exactly.

As stated, this model can be trained entirely with non-parallel data with maximum likelihood. To achieve voice conversion, the target features are generated according to
\begin{equation}
  \label{eqn:gmmmapping}
  \hat{\bm{y}}_t = \sum_{i=1}^KP(\bm{z}_t = i|\bm{x}_t)\left[\bm{\nu}_i +\bm{\Gamma}_i\bm{\Sigma}^{-1}_i(\bm{x}_t - \bm{\mu}_i)\right]
\end{equation}
The parameters $\bm{\nu}_i$ and $\bm{\Gamma}_i$ must be learned using data consisting of $(\bm{x}_t, \bm{y}_t)$ pairs. This is done by minimizing the mean square error (MSE) $||\bm{y}_t-\hat{\bm{y}}_t||^2$ on the training data. After training, conversion proceeds by inferring $P(\bm{z}_t = i|\bm{x}_t)$ from the source features $\{\bm{x}_t\}$ and generating the converted features $\{\hat{\bm{y}}_t\}$ with Eq. \ref{eqn:gmmmapping}.

This model treats conversion of each frame independently, and so the converted speech does not always mimic the dynamics of real speech. Later, this approach was extended to include temporal information relating to the trajectory of features \cite{toda2007voice, takamichi2015modulation}.

\subsection{Neural network VC with parallel data}
Motivated by the desire to move VC systems beyond the shallow conversion offered by the GMM based systems, there have been an increasing number of attempts at applying neural networks to the problem \cite{desai2009voice, sun2015voice, nakashika2013voice, kaneko2017sequence}. Many methods treat VC as a purely supervised learning problem, in which some hidden features $\bm{z}_t$ are inferred from the source input by the initial layers of the neural network, and then transformed by the final layers into the target output. Later methods \cite{sun2015voice, kaneko2017sequence} build on this by exploiting the flexibility of recurrent and convolutional neural networks to model the temporal aspects of the input, resulting in better conversion and less sensitivity to alignment errors in the training data. A further improvement made in \cite{kaneko2017sequence} replaces the usual cost function with an adversarially learned similarity metric that does away with the buzzing introduced by training with MSE.

While these methods have been successful when trained on datasets consisting of a hundred or more parallel utterances, performance degrades when less data is available. One way to avoid this problem is to consider a combined approach to increase the efficiency that the parallel data is used by augmenting the training with additional unpaired utterances. This combined approach has been relatively unexplored in the neural network VC literature. One of the few works \cite{mohammadi2014voice} that addresses this possibility carries out unsupervised pre-training of a deep autoencoder with data from multiple speakers, followed by a fine tuning step using parallel data. This is reminiscent of the unsupervised GMM training followed by the supervised learning of the conversion parameters discussed above. 

Further work on this autoencoding approach \cite{mohammadi2015semi} showed that pre-training separate autoencoders for the source and target (a process that can be done without parallel data), and then fine tuning the source encoder and target decoder using parallel data improves the source to target voice conversion. Conversion using this method can be carried out by obtaining a latent $\bm{z}_t$ from the source encoder and then decoding it into the target features by the target decoder. While this work clearly shows the benefits of semi-supervised training, to our knowledge the possibility of simultaneous training with a single objective remains unexplored.

\section{Proposed semi-supervised method}
\label{sec:proposed}

In developing our semi-supervised approach, we wish to retain several desirable features from both the GMM VC systems and the parallel data neural network systems discussed in Sec. \ref{sec:related}.  In the GMM systems, we note that the model parameters involved in computing the latent variables $\mathbf{z}_t$ do not require parallel training data to be learned, which decreases the amount of information that must be obtained via supervision.  Additionally, we note that there is an efficient (in this case exact) inference procedure for obtaining $\mathbf{z}_t$ given the input features that is vital for computationally tractable training and inference.  

The neural network systems also have an efficient procedure for computing $\mathbf{z}_t$, however it is qualitatively different from the probabilistic procedure in the GMMs owing to the highly nonlinear nature of the networks.  By virtue of this more flexible inference procedure, the neural network models are better able to handle the sequential nature of the input features, and model more complex relationships between the latent variables and the source/target features.  

Therefore, in constructing the semi-supervised approach, we seek a method that 1) can learn some model parameters at least partially from non-parallel data as in the GMM VC methods, 2) has an efficient and well defined probabilistic procedure for obtaining latent variables $\mathbf{z}_t$ as in the GMM methods, 3) can model complex relationships between the latent variables and source/target features as can the prior neural network models, and 4) can flexibly model the long sequential nature of the input as in the neural network approaches.

To do this we assume a latent variable sequence $\bm{Z}=\{z_1, z_2,...,z_T\}$ generates both the source and target sequences, $\bm{X} = \{x_1, x_2, ..., x_T\}$ and $\bm{Y} = \{y_1, y_2, ..., y_T\}$ respectively. We do not treat each frame $t=\{1, 2, ..., T\}$ independently, so each $x_t$, $y_t$ depends on the entire sequence of latent variables. We model the conditional distributions with factorized Gaussians:
\begin{equation}
  \begin{split}
    p\left(\bm{X}|\bm{Z}\right) &= \prod_{t=1}^T\mathcal{N}\left(x_t|f^{(x)}_{\theta_x}\left(\bm{Z},t\right),\sigma^2\cdot\mathbb{I}\right)\\
    p\left(\bm{Y}|\bm{Z}\right) &= \prod_{t=1}^T\mathcal{N}\left(y_t|f^{(y)}_{\theta_y}\left(\bm{Z},t\right),\sigma^2\cdot\mathbb{I}\right)\\
    p\left(\bm{Z}\right) &= \prod_{t=1}^T\mathcal{N}\left(z_t|0,\mathbb{I}\right)
  \end{split}
\end{equation}

Here, $\mathcal{N}\left(\cdot|f_\theta(\cdot),\sigma^2\cdot\mathbb{I}\right)$ is a multivariate normal distribution with mean given by the function $f_\theta\left(\cdot\right)$, which depends on parameters $\theta$, with a diagonal covariance matrix with nonzero elements set to be $\sigma^2$ for simplicity. $f^{(x)}_{\theta_x}\left(\bm{Z},t\right)$ and $f^{(y)}_{\theta_y}\left(\bm{Z},t\right)$ are separate functions for the source and target speakers respectively that capture the dependence of the source and target features on the sequence of latent variables. 

Exact inference is prohibitively expensive in this model, owing to both the nonlinearity of $f^{(x)}_{\theta_x}\left(\bm{Z},t\right)$ and $f^{(y)}_{\theta_y}\left(\bm{Z},t\right)$, and the large number of parent nodes for each $x_t$, $y_t$. However, approximate inference can be carried out in the variational autoencoder framework \cite{DBLP:journals/corr/KingmaW13}, which has shown success in semi-supervised classification problems \cite{kingma2014semi}. This requires the use of an approximate inference model such that the problem of finding $p\left(\bm{Z}|\bm{X}\right)$ or $p\left(\bm{Z}|\bm{Y}\right)$ is replaced with the approximation $q\left(\bm{Z}|\cdot\right)\approx p\left(\bm{Z}|\cdot\right)$, where $q\left(\bm{Z}|\cdot\right)$ is defined as

\begin{equation}
    q(\bm{Z}|\cdot) = \prod_{t=1}^T\mathcal{N}\left(z_t|g^{(\mu)}_{\phi_\mu}(\cdot,t),g^{(\sigma^2)}_{\phi_\sigma}(\cdot,t)\cdot\mathbb{I}\right)\\
\end{equation}

Here, $g^{(\mu)}_{\phi_\mu}(\cdot,t)$ is a function parameterized by $\phi_\mu$ that represents the mean of the multivariate normal, $g^{(\sigma^2)}_{\phi_\sigma}(\cdot,t)$ represents the diagonal elements of the covariance matrix. Note that the same function is used for both $\bm{X}$ and $\bm{Y}$. This choice was made with the intent that $\bm{Z}$ should be shared for both speakers. The functions $f^{(x)}_{\theta_x}\left(\bm{Z},t\right)$, $f^{(y)}_{\theta_y}\left(\bm{Z},t\right)$, $g^{(\mu)}_{\phi_\mu}(\cdot,t)$ and $g^{(\sigma^2)}_{\phi_\sigma}(\cdot,t)$ complete the specification of the model, and may be approximated with neural networks.

\subsection{Training objective}
Ideally, the parameters $\theta_x, \theta_y, \phi_\mu, \phi_\sigma$ would be learned via maximum likelihood. However, in this model exact likelihood calculations are prohibitive, and so we maximize a lower bound on the log-likelihood \cite{DBLP:journals/corr/KingmaW13} $\log\left(p\left(\bm{X}, \bm{Y}\right)\right)\geq\mathcal{L}_{X,Y}$ where 
\begin{equation}
  \label{eqn:bound}
  \begin{split}
  \mathcal{L}_{X,Y}&=
  \frac{1}{2\sigma^2}\sum_{t=1}^T\mathbb{E}_{z_t\sim q(\bm{Z}|\cdot)}\left(-||x_t -f^{(x)}_{\theta_x}\left(\bm{Z},t\right)||^2 \right) + \\
  &\frac{1}{2\sigma^2}\sum_{t=1}^T\mathbb{E}_{z_t\sim q(\bm{Z}|\cdot)}\left(-||y_t -f^{(y)}_{\theta_y}\left(\bm{Z},t\right)||^2 \right) +\\
  &-\mathcal{D}_{KL}\left(q(\bm{Z}|\cdot)||p\left(\bm{Z}\right)\right)
  \end{split}
\end{equation}
For semi-supervised training, we must consider the case where both $\bm{X},\bm{Y}$ are known, the case were $\bm{X}$ is known but $\bm{Y}$ is not, and the case where $\bm{Y}$ is known but $\bm{X}$ is not. In the first case, the bound on the log-likelihood is given by Eq. \ref{eqn:bound}. In the case where only $\bm{X}$ is known, the expectation in Eq. \ref{eqn:bound} involving $\bm{Y}$ is constant owing to the form of $p\left(\bm{Y}|\bm{Z}\right)$, and so we want to maximize $\log\left(p\left(\bm{X}\right)\right)\geq\mathcal{L}_{X,\_}$ where (after dropping the constant terms)
\begin{equation}
  \begin{split}
  \mathcal{L}_{X,\_}&=
  \frac{1}{2\sigma^2}\sum_{t=1}^T\mathbb{E}_{z_t\sim q(\bm{Z}|\bm{X})}\left(-||x_t -f^{(x)}_{\theta_x}\left(\bm{Z},t\right)||^2 \right) + \\
  &-\mathcal{D}_{KL}\left(q(\bm{Z}|\bm{X})||p\left(\bm{Z}\right)\right)
  \end{split}
\end{equation}
Equivalently, when $\bm{Y}$ is known but $\bm{X}$ is not, we maximize $\log\left(p\left(\bm{Y}\right)\right)\geq\mathcal{L}_{\_,Y}$, which has the same form as $\mathcal{L}_{X,\_}$ but has $\bm{Y}$ in place of $\bm{X}$.

The bound on the entire dataset is therefore
\begin{equation}
  \label{eqn:costfxn}
  \mathcal{L}=\sum_{X\in\{X,\_\}}\mathcal{L}_{X,\_}+\sum_{Y\in\{\_,Y\}}\mathcal{L}_{\_,Y} + \sum_{X,Y\in\{X,Y\}}\mathcal{L}_{X,Y}
\end{equation}
In practice, we compute the expectations in $\mathcal{L}_{X,\_}, \mathcal{L}_{\_,Y}$ using a single sample. For the expectations in $\mathcal{L}_{X,Y}$ we would ideally use an approximation of $p(\bm{Z}|\bm{X}, \bm{Y})$. However, in the case where parallel training data is limited, directly learning such an approximation is prohibitively complex. Instead, we use two samples to compute this quantity, one using $q(\bm{Z}|\bm{X})$ and another using $q(\bm{Z}|\bm{Y})$. 

\subsection{Baseline systems and proposed modifications}
As a baseline approach, we implemented the DBLSTM model from \cite{sun2015voice}. This model consists of four bidirectional long short-term memory (BLSTM) layers of sizes 128, 256, 256, 128 each. With the exception of using the WORLD vocoder instead of the STRAIGHT vocoder \cite{kawahara1999restructuring}, we use this model as described in the original work. 

We then extend the DBLSTM model so that it can be trained with semi-supervision. To do this, we interpret the first two BLSTM layers of the model as the encoder portion and apply two separate affine transformations to the 256 dimensional output for each time frame to compute $g^{(\mu)}_{\phi_\mu}(\cdot,t)$ and $\log\left(g^{(\sigma^2)}_{\phi_\sigma}(\cdot,t)\right)$ respectively. In this way, we obtain the distribution $q(\bm{Z}|\cdot)$ describing a 256 dimensional latent variable $\bm{Z}$. This $\bm{Z}$ acts as the inputs to the final two layers of the DBLSTM model which we interpret as the decoder portion. We have two separate decoder portions (each using $\sigma^2=10^{-3}$) with different parameters but an otherwise identical architecture such that one acts as $f^{(x)}_{\theta_x}\left(\bm{Z},t\right)$ and the other as $f^{(y)}_{\theta_y}\left(\bm{Z},t\right)$. The resulting model can be trained with the semi-supervised cost function described in Eq. \ref{eqn:costfxn}.

To verify that this architectural modification by itself does not significantly impact performance in the absence of semi-supervision, we also carried out experiments on the modified model with purely supervised training. This corresponds to only optimizing the $\mathcal{L}_{X,Y}$ term of Eq. \ref{eqn:costfxn}. For approximate inference in this model, we use $\bm{X}$.  We call this model the DBLSTM+VAE, to distinguish it from the baseline DBSLTM and denote that it has been reinterpreted in the variational autoencoder (VAE) framework. 
\section{Experiments}
\label{sec:experiments}

\begin{figure}[t!]
\centering
\begin{minipage}[t]{0.45\textwidth}
  \includegraphics[width=1\linewidth]{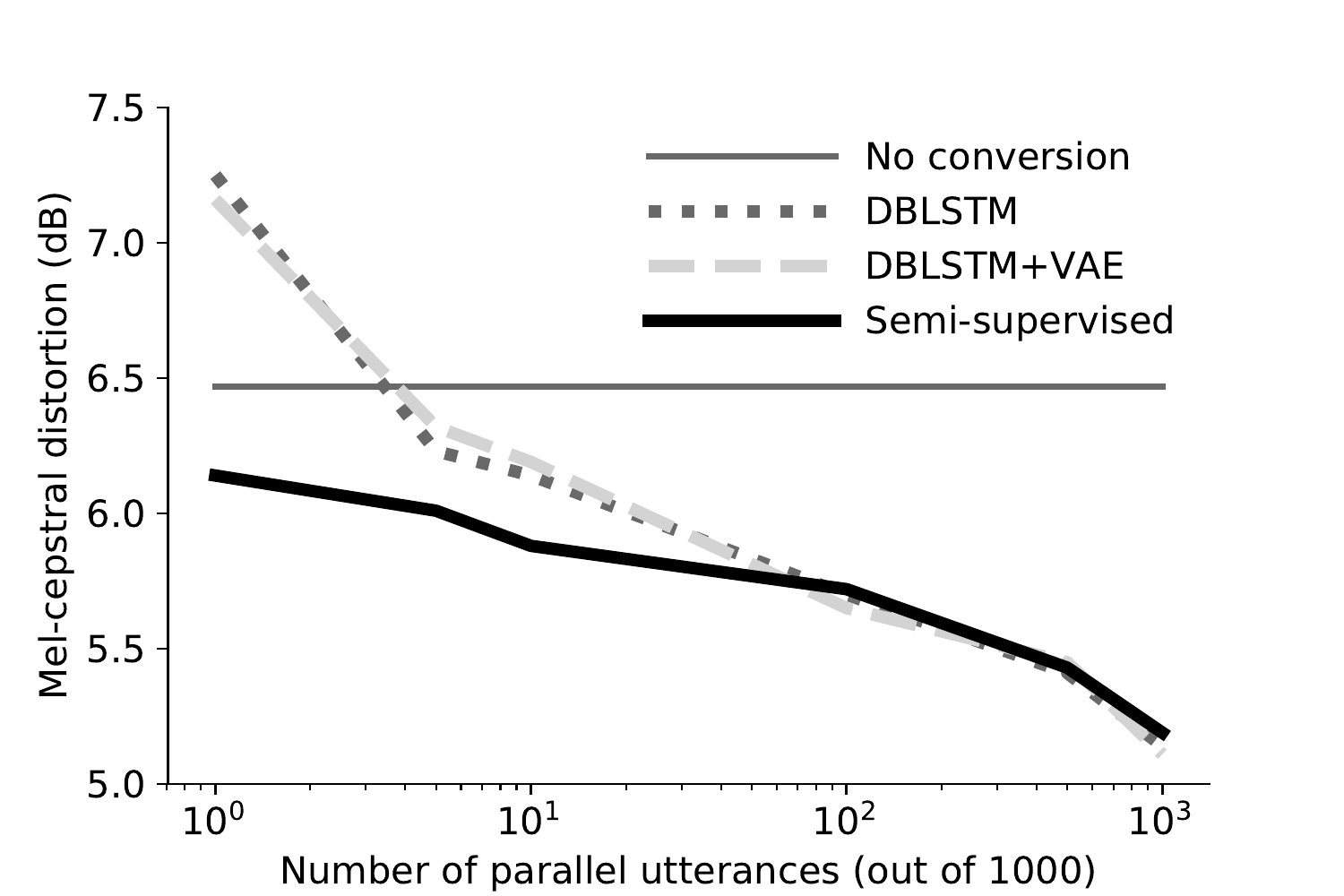}
  \caption{\ninept Performance of semi-supervised training vs. fully supervised training for a varying number of parallel training utterances out of a total of 1000 utterances.}
  \label{fig:parallelsweep}
\end{minipage}
\hfill
\end{figure}

To evaluate the performance of our semi-supervised method, we carried out experiments with varying amounts of parallel and non-parallel training data.  To create the different datasets, we drew samples from the CLB$\to$SLT (both females) pair in the CMU Arctic corpus \cite{kominek2003cmu}.  To generate the parallel data corpus, we drew paired $\mathbf{X}, \mathbf{Y}$ samples at random from both the A and B partitions of the CMU Arctic, and time aligned the target features to the source features using dynamic time warping (DTW).  For the non-parallel $\mathbf{X}, \_$ and $\_, \mathbf{Y}$ datasets, we drew samples at random from the remaining unselected samples in the A and B partitions respectively, ensuring that the same prompt does not appear in both the $\mathbf{X}, \_$ and $\_, \mathbf{Y}$ datasets.  We considered datasets consisting of at most 1000 utterances, and so the remaining 93 utterances from the A partition were used as a validation dataset, while testing was carried out on the remaining 39 utterances of the B partition.

In line with \cite{sun2015voice}, we extracted 50 Mel-cepstral coefficients (MCEPs) from the spectral envelope obtained using the WORLD vocoder \cite{morise2016world} as features along with fundamental frequency contours (F$_0$) and aperiodicities (APs). We used $16$kHz audio, an FFT size of $1024$ with a hop length of $5$ms.  The zeroth cepstral coefficient was left unmodified, and the remaining 49 coefficients were Gaussian normalized and used as the features for VC. The F$_0$ contours were converted via the usual log Gaussian normalization \cite{liu2007high}, and the APs were used directly from the input without modification.

To obtain an objective performance measure, we evaluated each model using mel-cepstral distortion \cite{kubichek1993mel, kominek2008synthesizer} (MCD). We measured the average MCD between paired source and target utterances before conversion to be $6.47$ dB. While the proposed method treats the source and target symmetrically, we only carried out the evaluation for source to target conversion to directly compare to the supervised approaches which treat the problem asymmetrically. 

\subsection{Increasing amounts of parallel data}
\label{subsec:experiment1}
To verify the effectiveness of the semi-supervised training under realistic constraints on the amount of parallel training data, we considered a fixed training data budget of $N=1000$ total utterances (roughly 1 hour from each speaker), and varied the number of parallel training utterances.  The non-parallel utterances were evenly split between the source and target.

Results for varying amounts of parallel training data are shown in Fig. \ref{fig:parallelsweep}. When only a small fraction of the training data consists of parallel utterances, we find that training with semi-supervision gives far better performance than purely supervised training. We also see that for datasets that contain mostly parallel data, the proposed semi-supervised method gives equivalent performance to the fully supervised approach as expected.  For intermediate amounts of parallel and non-parallel data, the semi-supervised training smoothly interpolates between these two cases, consistently performing better than or equal to the purely supervised training.

\begin{table}[t]
  \small
  \centering
  \caption{\ninept Mean Opinion Score (MOS) of models trained on 1, 10, and 1000 parallel utterances, out of a total of 1000 utterances. Error bars represent 95\% confidence intervals.}
  \label{tab:mostest}
  \begin{tabular}{llll}
    \toprule
    \textbf{ }      & \textbf{1} & \textbf{10} & \textbf{1000}          \\
    \midrule
    DBLSTM\cite{sun2015voice} & 1.64$\pm$0.13 & 2.49$\pm$0.17 & 3.39$\pm$0.16 \\
    DBLSTM + VAE              & 1.73$\pm$0.14 & 2.56$\pm$0.14 & 3.40$\pm$0.14 \\
    Semi-Supervised           & \textbf{2.93$\pm$0.16} & \textbf{2.99$\pm$0.16} & \textbf{3.63$\pm$0.16} \\
    \bottomrule
  \end{tabular}
\end{table}
To verify that the audio quality remains high as the amount of parallel training data is varied, we also carried out a subjective evaluation of the quality of the converted audio from algorithms trained on 1, 10, and 1000 parallel utterances. We evaluated Mean Opinion Score (MOS) using 40 listeners on Amazon Mechanical Turk, over 90\% of which were native English speakers. Each listener was asked to rate the quality of 3 utterances from each model on a 5-point scale (1=Bad, 2=Poor, 3=Fair, 4=Good, 5=Excellent). Results of this evaluation are shown in Table \ref{tab:mostest}. We see that the semi-supervised training gives much higher quality audio when only a small amount of parallel data is available. Measuring both conversion (via MCD) and quality (with MOS), we find that when only a small amount of parallel data is available the semi-supervised approach achieves voice conversion of quality comparable to the supervised approaches trained with a much larger parallel dataset.

\subsection{Increasing amounts of non-parallel data}
\label{subsec:experiment2}

\begin{figure}[t!]
\centering
\begin{minipage}[t]{0.45\textwidth}
  \centering
  \includegraphics[width=1\linewidth]{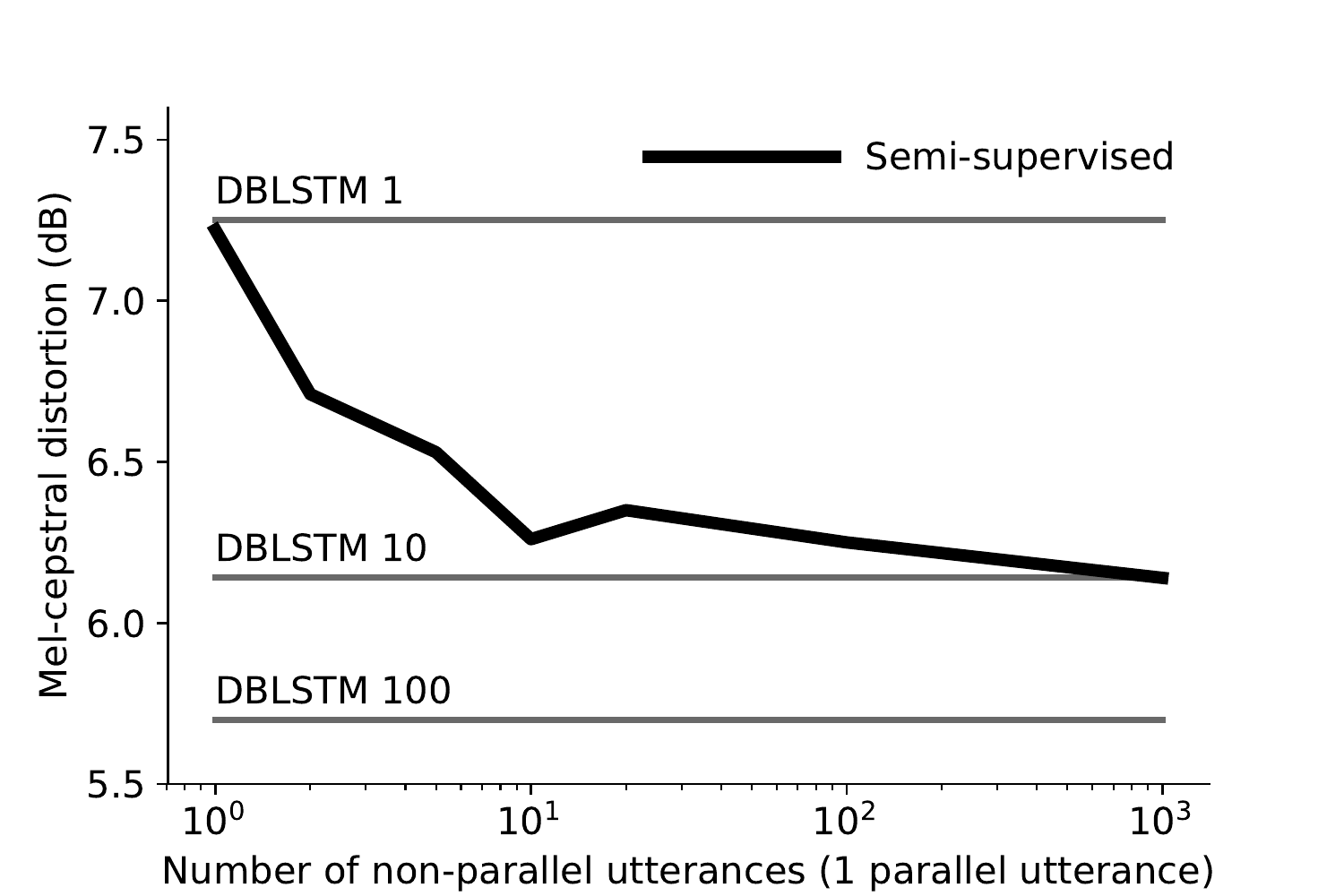}
  \caption{\ninept Performance of semi-supervised method with a single parallel utterance and an increasing number of non-parallel utterances. Horizontal lines show performance of supervised approach with a varying number of parallel utterances.}
  \label{fig:nonparallelsweep}
\end{minipage}
\hfill
\end{figure}

As the results of Fig. \ref{fig:parallelsweep} and Table \ref{tab:mostest} show, the inclusion of non-parallel data in the semi-supervised training leads to improved performance, and performance of all methods continues to improve for larger amounts of parallel training data.  However, in the semi-supervised case, it may also be possible to improve performance by increasing the amount of non-parallel data used in training.

To test this effect we considered a fixed data budget of only one parallel utterance, and varied the amount of non-parallel utterances used in the semi-supervised training. Results of this experiment are shown in Fig. \ref{fig:nonparallelsweep}. We find that increasing the amount of non-parallel data used in training improves the VC performance as was the case for increasing amounts of parallel data.  However the rate of improvement decreases for larger amounts of non-parallel data, while we did not observe this with larger amounts of parallel data.

This suggests that in creating a dataset for a semi-supervised VC system, a trade off must be made between gathering harder to obtain but more informative parallel training examples, and easier to obtain but less informative non-parallel examples.  While we do not investigate this here, it is an interesting and promising avenue of future work to devise methods of improving the efficiency in which non-parallel training data is used as this will determine the overall cost and difficulty of creating a VC training dataset.

\section{Conclusion}
\label{sec:conclusion}
We have proposed a new semi-supervised method for achieving voice conversion using both parallel and non-parallel data. This method incorporates both types of data simultaneously during training by optimizing a variational objective defined for paired and unpaired utterances. When only a small number of parallel utterances are available, we show that incorporating this method into an existing neural network model improves the accuracy and perceptual quality of the converted speech compared to supervised training. We also find that increasing the amount of non-parallel data continues to improve voice conversion.  This opens up the possibility of training VC systems with more flexible datasets consisting of mixed parallel and non-parallel data.
\clearpage
\bibliographystyle{IEEEtran}
\bibliography{cites}

\end{document}